# An Application of Reinforcement Learning to Dialogue Strategy Selection in a Spoken Dialogue System for Email

**Marilyn A. Walker**　　　　　　　　　　　　　　　　　　　WALKER@RESEARCH.ATT.COM
*AT&T Shannon Laboratory*
*180 Park Ave., Bldg 103, Room E103*
*Florham Park, NJ 07932*

## Abstract

This paper describes a novel method by which a spoken dialogue system can learn to choose an optimal dialogue strategy from its experience interacting with human users. The method is based on a combination of reinforcement learning and performance modeling of spoken dialogue systems. The reinforcement learning component applies Q-learning (Watkins, 1989), while the performance modeling component applies the PARADISE evaluation framework (Walker et al., 1997) to learn the performance function (reward) used in reinforcement learning. We illustrate the method with a spoken dialogue system named ELVIS (EmaiL Voice Interactive System), that supports access to email over the phone. We conduct a set of experiments for training an optimal dialogue strategy on a corpus of 219 dialogues in which human users interact with ELVIS over the phone. We then test that strategy on a corpus of 18 dialogues. We show that ELVIS can learn to optimize its strategy selection for agent initiative, for reading messages, and for summarizing email folders.

## 1. Introduction

In the past several years, it has become possible to build spoken dialogue systems that can communicate with humans over the telephone in real time. Systems exist for tasks such as finding a good restaurant nearby, reading your email, perusing the classified advertisements about cars for sale, or making travel arrangements (Seneff, Zue, Polifroni, Pao, Hetherington, Goddeau, & Glass, 1995; Baggia, Castagneri, & Danieli, 1998; Sanderman, Sturm, den Os, Boves, & Cremers, 1998; Walker, Fromer, & Narayanan, 1998). These systems are some of the few realized examples of real time, goal-oriented interactions between humans and computers. Yet in spite of 30 years of research on algorithms for dialogue management in task-oriented dialogue systems, (Carbonell, 1971; Winograd, 1972; Simmons & Slocum, 1975; Bruce, 1975; Power, 1974; Walker, 1978; Allen, 1979; Cohen, 1978; Pollack, Hirschberg, & Webber, 1982; Grosz, 1983; Woods, 1984; Finin, Joshi, & Webber, 1986; Carberry, 1989; Moore & Paris, 1989; Smith & Hipp, 1994; Kamm, 1995) *inter alia*, the design of the dialogue manager in real-time, implemented systems is still more of an art than a science (Sparck-Jones & Galliers, 1996). This paper describes a novel method, and experiments that validate the method, by which a spoken dialogue system can *learn* from its experience with human users to optimize its choice of dialogue strategy.

　　The dialogue manager of a spoken dialogue system processes the user's utterance and then chooses in real time *what* information to communicate to the human user and *how* to communicate it. The choice it makes is called its *strategy*. The dialogue manager can be naturally formulated as a state machine, where the state of the dialogue is defined by a set of state variables representing observations of the user's conversational behavior, the results of accessing various information databases, and aspects of the dialogue history. Transitions between states are driven by the system's dialogue strategy. In a typical system, there are a large number of potential strategy choices at each state of a dialogue.

For example, consider one choice faced by ELVIS (EmaiL Voice Interactive System) a spoken dialogue system that supports access to a user's email by phone. ELVIS provides verbal summaries of a user's email folders, but there are many ways to summarize a folder (Sparck-Jones, 1993, 1999). A summary could consist of a simple statement about the number of messages in different folders, e.g., *You have 5 new messages*, or it could provide much more detail about the messages in a particular folder, e.g., *In your messages from Kim, you have one message about a meeting, and a second about interviewing Antonio*. ELVIS must decide which of many properties of a message to mention, such as the message's status, its sender, or the subject of the message.[1]

Decision theoretic planning can be applied to the problem of choosing among dialogue strategies, by associating a utility $U$ with each strategy (action) choice and by positing that spoken dialogue systems should adhere to the Maximum Expected Utility Principle (Keeney & Raiffa, 1976; Russell & Norvig, 1995),

> **Maximum Expected Utility Principle**: An optimal action is one that maximizes the expected utility of outcome states.

Thus, ELVIS can act optimally by choosing a strategy $a$ in state $S_i$ that maximizes $U(S_i)$. This formulation however simply leaves us with the problem of how to derive the utility values $U(S_i)$ for each dialogue state $S_i$. Several reinforcement learning algorithms based on dynamic programming specify a way to calculate $U(S_i)$ in terms of the utility of a successor state $S_j$ (Bellman, 1957; Watkins, 1989; Sutton, 1991; Barto, Bradtke, & Singh, 1995), so if the utility for the final state of the dialogue were known, it would be possible to calculate the utilities for all the earlier states, and thus determine a policy which selects only optimal dialogue strategies.

Previous work suggested that it should be possible to treat dialogue strategy selection as a stochastic optimization problem in this way (Walker, 1993; Biermann & Long, 1996; Levin, Pieraccini, & Eckert, 1997; Mellish, Knott, Oberlander, & O'Donnell, 1998). However in (Walker, 1993), we argued that the lack of a performance function for assigning a utility to the final state of a dialogue was a critical methodological limitation. There seemed to be three main possibilities for a simple reward function: user satisfaction, task completion, or some measure of user effort such as elapsed time for the dialogue or the number of user turns. But it appeared that any of these simple reward functions on their own fail to capture essential aspects of the system's performance. For example, the level of user effort to complete a dialogue task is system, domain and task dependent. Moreover, high levels of effort, e.g., the requirement that users confirm the system's understanding of each utterance, do not necessarily lead to concomitant increases in task completion, but do lead to significant decreases in user satisfaction (Shriberg, Wade, & Price, 1992; Danieli & Gerbino, 1995; Kamm, 1995; Baggia et al., 1998). Furthermore, user satisfaction alone fails to reflect the fact that the system will not be successful unless it helps the user complete a task. We concluded that the relationship between these measures is both interesting and complex and that a method for deriving an appropriate performance function was a necessary precursor to applying stochastic optimization algorithms to spoken dialogue systems. In (Walker, Litman, Kamm, & Abella, 1997a), we proposed the PARADISE method for learning a performance function from a corpus of human-computer dialogues.

---

1. All of the strategies implemented in ELVIS are summarized in Figure 1. Note that due to practical constraints, we have only implemented strategy choices in a subset of states, and that ELVIS uses a fixed strategy in other states. In Section 2, we describe in detail the strategy choices that ELVIS explores in addition to choices about summarization, namely choices among strategies for controlling the dialogue initiative and for reading multiple messages.

In this work, we apply the PARADISE model to learn a performance function for ELVIS, which we then use for calculating the utility of the final state of a dialogue in experiments applying reinforcement learning to ELVIS's selection of dialogue strategies. Section 2 describes the implementation of a version of ELVIS that randomly explores alternate strategies for initiative, for reading messages, and for summarizing email folders. Section 3 describes the experimental design in which we first use this exploratory version of ELVIS to collect a training corpus of conversations with 73 human users carrying out a set of three email tasks. Section 4 describes how we apply reinforcement learning to the corpus of 219 dialogues to optimize ELVIS's dialogue strategy decisions. We then test the optimized policy in an experiment in which six new users interact with ELVIS to complete the same set of tasks, and show that the learned policy performs significantly better than the exploratory policy used during the training phase.

## 2. ELVIS Spoken Dialogue System

We started the process of designing ELVIS by conducting a Wizard-of-Oz experiment in which we recorded dialogues with six people accessing their email remotely by talking to a human who was playing the part of the spoken dialogue system. The purpose of this experiment was to identify the basic functionality that should be implemented in ELVIS. The analysis of the resulting dialogues suggested that ELVIS needed to support content-based access to email messages by specification of the subject or the sender field, verbal summaries of email folders, reading the body of an email message, and requests for help and repetition of messages (Walker et al., 1997b, 1998).

Given these requirements, we then implemented ELVIS using a general-purpose platform for spoken dialogue systems (Kamm et al., 1997). The platform consists of a dialogue manager (described in detail in Section 2.2), a speech recognizer, an audio server for both voice recordings and text-to-speech (TTS), an interface between the computer running ELVIS and the telephone network, and modules for specifying the rules for spoken language understanding and application specific functions.

The speech recognizer is a speaker-independent Hidden Markov Model (HMM) system, with context-dependent phone models for telephone speech, and constrained grammars defining the vocabulary that is permitted at any point in a dialogue (Rabiner, Juang, & Lee, 1996). The platform supports barge-in, so that the user can interrupt the system; barge-in is very important for this application so that the user can interrupt the system when it is reading a long email message.

The audio server can switch between voice recordings and text-to-speech (TTS) and integrate voice recordings with TTS. The TTS technology is concatenative diphone synthe-sis (Sproat & Olive, 1995). ELVIS uses only TTS since it would not be possible to pre-record, and then concatenate, all the words necessary for realizing the content of email messages.

The spoken language understanding (SLU) module consists of a set of rules for specifying the vocabulary and allowable utterances, and an associated set of rules for translating the user's utterance into a domain-specific semantic representation of its meaning. The syntactic rules are converted into an FSM network that is used directly by the speech recognizer (Mohri, Pereira, & Riley, 1998). The semantic rule that is associated with each syntactic rule maps the user's utterance directly to an application specific template consisting of an application function name and its arguments. These templates are then converted directly to application specific function calls specified in the application module. The understanding module also supports dynamic grammar generation and loading because the recognizer vocabulary must change during the interaction, e.g., to support selection of email messages by content fields such as sender and subject.

The application module provides application specific functions, e.g., functions for accessing message attributes such as subject and sender, and functions for making these realizable as speech so that they can be used to instantiate variables in spoken language generation.

### 2.1 ELVIS's Dialogue Manager and Strategies

ELVIS's dialogue manager is based on a state machine where one or more dialogue strategies can be explored in each state. The *state* of the dialogue manager is defined by a set of state variables representing various items of information that the dialogue manager uses in deciding what to do next. The state variables encode various observations of the user's conversational behavior, such as the results of processing the user's speech with the spoken language understanding (SLU) module, and results from accessing information databases relevant to the application, as well as certain aspects of the dialogue history. A *dialogue strategy* is a specification of what the system should say; in ELVIS this is represented as a template with variables that must be instantiated by the current context. In some states the system always executes the same dialogue strategy and in other states alternate strategies are explored. All of the strategies implemented in ELVIS are summarized in Figure 1. A complete specification of which dialogue strategy should be executed in each state is called a *policy* for a dialogue system.

To develop a version of ELVIS that supported exploring a number of possible policies, we implemented several different choices in particular states of the system. Our goal was to implement strategy choices in states where the optimal strategy was not obvious a priori. For the purpose of illustrating the dialogue strategies we explored, consider a situation in which the user is attempting to execute the following task (one of the tasks used in the experimental data collection described in Section 3):

- Task 1.1: You are working at home in the morning and plan to go directly to a meeting when you go into work. Kim said she would send you a message telling you where and when the meeting is. Find out **the Meeting Time** and **the Meeting Place**.

To complete this task, the user needs to find a message from Kim about a meeting in her inbox and listen to it. There are many possible strategies that ELVIS could use to help the user accomplish this task. Below, we first describe the dialogue strategies from Figure 1 that ELVIS makes choices among, then describe in detail the complete state machine, the dialogue strategies from Figure 1 that are used in states where there is no choice among dialogue strategies, and the space of policies that ELVIS can execute. We provide several detailed examples of dialogues that can be generated by ELVIS's dialogue manager's state machine.

| Strategy Type | Choices Explored? | Strategy Choices |
|---|---|---|
| Initiative | yes | System-Initiative (SI), Mixed-Initiative (MI) |
| Summarization | yes | SummarizeBoth (SB), SummarizeSystem (SS), SummarizeChoicePrompt (SCP) |
| Reading | yes | Read-First (RF), Read-Summary-Only (RSO), Read-Choice-Prompt (RCP) |
| Request-Info | no | AskUserName, Ask-Which-Selection (AskWS), Ask-Selection-Criteria (AskSC), |
| Provide-Info | no | Read-Message |
| Help | no | AskUserName-Help, SI-Top-Help, MI-Top-Help, Read-Message-Help, AskWS-Help, AskSC-Help |
| Timeout | no | AskUserName-Timeout, Read-Timeout, SI-Top-Timeout, MI-Top-Timeout, Read-Message-Timeout, AskWS-Timeout, AskSC-Timeout |
| Rejection | no | AskUserName-Reject, SI-Top-Reject, MI-Top-Reject, AskWS-Reject, AskSC-Reject, Read-Message-Reject |

Figure 1: ELVIS's Dialogue Strategies. ELVIS explores choices in Initiative, Summarization and Read Strategies and uses fixed strategies elsewhere.

### 2.1.1 STRATEGY CHOICES IN ELVIS

One strategy choice in our experiments involved the exploration of two different strategies for managing the dialogue initiative: system-initiative and mixed-initiative. In the implementation of ELVIS tested here, the choice of initiative was made early in the dialogue and the same style of initiative was then used throughout the dialogue whenever there could have been a choice between the two in order to avoid confusing the user.

Previous research suggested that the system-initiative strategy in Figure 2 in which ELVIS is directive with the user might be the best strategy, given the current state of the art in speech recognition (Kamm, 1995). In Figure 2, each of ELVIS's utterances is labelled with the name of the dialogue strategy that ELVIS is using to produce the utterance, e.g., utterance E1 is labelled SI-Top, the name for the system-initiative top level strategy. The choice of the system-initiative strategy also affects utterances E2 and E3; in these utterances ELVIS continues the dialogue by requesting particular items of information. In general, the system-initiative strategy constrains what the user can say at each point in the dialogue, by requesting particular items of information one at a time, and the user is constrained to only providing the information that has been asked for. The reason why this strategy is recommended given the current state of the art in speech recognition is that it guides the user into saying things that the system is more likely to understand, and the recognizer vocabulary at each state is constrained to responses to the question that is asked. This strategy might be dispreferred by users however, because it could be tedious for the user to be constrained to providing one item of information at a time. In the dialogue in Figure 2, it takes four user utterances to access the message from Kim needed to complete Task 1.1.

| | |
|---|---|
| E1: Hi, ELVIS here...You have 5 new or unread messages in your inbox. Say Read, Summarize, or say Help for more options. | SI-Top |
| U1: Read. | |
| E2: Select by Sender, Subject or in order of arrival? | Ask-Which-Selection |
| U2: Sender. | |
| E3: Which Sender? | Ask-Selection-Criterion |
| U3: Kim. | |
| E4: In the messages from Kim, there's 1 message about "Interviewing Antonio" and 1 message about "Meeting Today." The first message is titled, "Interviewing Antonio." It says, "I'd like to interview him. I could also go along to lunch. Kim." | Read-First (RF) |
| U4: Next. | |
| E5: This is a message about meeting today. It says the meeting this morning is at 10:00 in 2d516. See you there, Kim. | Read-Message |
| U5: I'm done here. | |

Figure 2: System-Initiative (SI) Dialogue Strategy: ELVIS explicitly prompts the user with what to say at each point of the dialogue. ELVIS's utterances are labelled with the strategy names.

Other research suggested that a better strategy is a mixed-initiative strategy that allows the user to take the initiative except when the user fails to say anything or the system is having trouble understanding the user (Smith & Hipp, 1994; Levin, Pieraccini, Eckert, Fabbrizio, & Narayanan, 1999; Webber & Joshi, 1982; Walker & Whittaker, 1990). The dialogue in Figure 3 illustrates the mixed-initiative dialogue strategy. Each of ELVIS's utterances in Figure 3 is labelled with the name of the dialogue strategy that ELVIS is using to produce the utterance, e.g., utterance E1 is labelled MI-Top, the name for the mixed-initiative top level strategy. Note that unlike the SI-Top strategy, the MI-Top strategy does not volunteer information and does not tell the user what she can say. Instead, the mixed-initiative strategy allows the user to take control of the dialogue and specify exactly what she wants to do in a single utterance, as in utterance U1 in Figure 3. However, if there is evidence that the user is having trouble or is confused, as when the user says nothing in utterance U2, the mixed-initiative strategy will take the initiative to ask the user for information or tell the user what she can say. In utterance E3 in Figure 3 the Read-Timeout strategy is triggered by the fact that the user's response was delayed longer than a pre-set timeout threshhold, and the system tells the user what actions are available at this point in the dialogue. In general, the mixed-initiative strategy can support more efficient dialogues; in the dialogue in Figure 2, it takes two user utterances to access the message from Kim needed to complete Task 1.1. However this efficiency gain may be offset by the increased likelihood of misunderstandings and user confusion.

| | |
|---|---|
| E1: Hi, ELVIS here. I've got your mail. | MI-Top |
| U1: Read me my messages from Kim. | |
| E2: In the messages from Kim, there's 1 message about "Interviewing Antonio" and 1 message about "Meeting Today." The first message is titled, "Interviewing Antonio." It says, "I'd like to interview him. I could also go along to lunch. Kim." | Read-First (RF) |
| U2: (*User says nothing within 5 seconds so the system takes the initiative*) | |
| E3: You can say repeat, to repeat the last message, or you can say next, to read the next message in this folder. You can say, I'm done here to exit this folder. | Read-Timeout |
| U3: Next message. | |
| E4: This is a message about meeting today. It says the meeting this morning is at 10:00 in 2d516. See you there, Kim. | ReadMessage |
| U4: I'm done here. | |

Figure 3: Mixed-Initiative (MI) Dialogue Strategy: ELVIS leaves it up to the user to take the initiative, unless the user seems to be having trouble with the system

| Summarize Strategy | Example Prompt |
|---|---|
| Summarize-Both (SB) | In your top level inbox, from Kim, there's 1 message about "Lunch." From Michael, there's 1 message about "Evaluation group meeting." From Noah, there's 1 message about "Call Me Tomorrow" and 1 message about "Interviewing Antonio." And from Owen, there's 1 message about "Agent Personality." |
| Summarize-System (SS) | In your top level inbox, there's 1 message from Kim, 2 messages from Noah, 1 message from Michael, and 1 message from Owen. |
| Summarize-Choice-Prompt (SCP) | E: Summarize by subject, by sender, or both? <br> U: Subject. <br> E: In your top level inbox, there's 1 message about "Lunch," 1 message about "Interviewing Antonio," 1 message about "Call Me Tomorrow," 1 message about "Evaluation Group Meeting," and 1 message about "Agent Personality." |

Figure 4: Alternate Summarization Strategies in response to a request to "Summarize my messages"

A different type of strategy choice involves ELVIS's decisions about how to present information to the user. We mentioned above that there are many different ways to summarize a set of items that the user wants information about. ELVIS explores the set of alternate summarization strategies illustrated in Figure 4; these strategies vary the message attributes that are included in a summary of the messages in the current folder. The Summarize-Both strategy (SB) uses both the sender and the subject attributes in the summary. When employing the Summarize-System strategy (SS), ELVIS summarizes by subject or by sender based on the current context. For instance, if the user is in the top level inbox, ELVIS will summarize by sender, but if the user is situated in a folder containing messages from a par-ticular sender, ELVIS will summarize by subject, as a summary by sender would provide no new information. The Summarize-Choice-Prompt (SCP) strategy asks the user to specify which of the relevant attributes to summarize by. See Figure 4.

Another type of information presentation choice occurs when a request from the user to read some subset of messages, e.g., *Read my messages from Kim*, results in multiple matching messages. The strategies explored in ELVIS are summarized in Figure 5. One choice is the Read-First strategy (RF) which involves summarizing all the messages from Kim, and then taking the initiative to read the first one. ELVIS used this read strategy in the dialogues in Figures 2 and 3. An alternate strategy for reading multiple matching messages is the Read-Summary-Only (RSO) strategy, where ELVIS provides information that allows users to refine their selection criteria. Another strategy for reading multiple messages is the Read-Choice-Prompt (RCP) strategy, where ELVIS explicitly tells the user what to say in order to refine the message selection criteria. See Figure 5.

| Read Strategy | Example Prompt |
|---|---|
| Read-First (RF) | In the messages from Kim, there's 1 message about "Interviewing Antonio" and 1 message about "Meeting Today." The first message is titled, "Interviewing Antonio." It says, "I'd like to interview him. I could also go along to lunch. Kim." |
| Read-Summary-Only (RSO) | In the messages from Kim, there's 1 message about "Interviewing Antonio" and 1 message about "Meeting Today." |
| Read-Choice-Prompt (RCP) | In the messages from Kim, there's 1 message about "Interviewing Antonio" and 1 message about "Meeting Today." To hear the messages, say, "Interviewing Antonio" or "Meeting." |

Figure 5: Alternate Read Strategies in response to a request to "Read my messages from Kim"

The remainder of ELVIS's dialogue strategies, as summarized in Figure 1, are fixed, i.e. multiple versions of these strategies are not explored in the experiments presented here.

### 2.1.2 ELVIS'S DIALOGUE STATE MACHINE

As mentioned above, a *dialogue strategy* is a choice the system makes, in a particular state, about what to say and how to say it. A *policy* for a dialogue system is a complete specification of which strategy to execute in each system state. A *state* is defined by a set of state variables. Ideally, the state representation corresponds to a dialogue model that summarizes the dialogue history compactly, but retains all the relevant information about the dialogue interaction so far. The notion of a dialogue model retaining all the relevant information is more formally known in reinforcement learning as a state representation that satisfies the *Markov Property*. A state representation satisfying the Markov Property is one in which the probability of being in a particular state $s$ with a particular reward $r$ after doing some action $a$ in a prior state can be estimated as a function of the action and the prior state, and not as a function of the complete dialogue history (Sutton & Barto, 1998). More precisely,

$$Pr(s_{t+1} = s', r_{t+1} = r | s_t, a_t) = Pr(s_{t+1} = s', r_{t+1} = r | s_t, a_t, r_t, s_{t-1}, a_{t-1}, r_{t-1}, \ldots R_1, s_0, a_0)$$

for all $s', r, s_t$ and $a_t$.

The Markov Property is guaranteed if the state representation encodes everything that the system has been able to observe about everything that happened in the dialogue so far. However, this representation would be too complex to estimate a model of the probability of various state transitions, and systems as complex as spoken dialogue system must in general utilize state representations which are as compact as possible.[2] However if the state representation is too impoverished, the system will lose too much relevant information to work well.

| Operations Variable | Abbrev | Possible Values |
|---|---|---|
| KnowUserName | (U) | 0,1 |
| InitStrat | (I) | 0,SI,MI |
| SummStrat | (S) | 0,SS,SCP,SB |
| ReadStrat | (R) | 0,RF,RSO,RCP |
| TaskProgress | (P) | 0,1,2 |
| CurrentUserGoal | (G) | 0,Read,Summarize |
| NumMatches | (M) | 0,1,N>1 |
| WhichSelection | (W) | 0,Sender (Snd),Subject (Sub),InOrder (InO) |
| KnowSelectionCriteria | (SC) | 0,1 |
| Confidence | (C) | 0,1 |
| Timeout | (T) | 0,1 |
| Help | (H) | 0,1 |
| Cancel | (L) | 0,1 |

Figure 6: Operations variables and possible values that define the operations vector for controlling all aspects of ELVIS's behavior. The abbreviations for the variable names and values are used as column headers for the Operations Variables in Figures 7, 8 and 9.

ELVIS's state space representation must obviously discriminate among states in which various strategy choices are explored, but in addition, there must be state variables to capture distinctions between a number of states in which ELVIS always executes the same strategy. The state variables that ELVIS keeps track of and their possible values are given in Figure 6. The KnowUserName (U) variable keeps track of whether ELVIS knows the user's name or not. The InitStrat (I), SummStrat (S) and ReadStrat (R) variables keep track of whether ELVIS has already employed a particular initiative strategy, summarize strategy or a reading strategy in the current dialogue, and if so, which strategy it has selected. This variable is needed because once ELVIS employs one of these strategies, that strategy is used consistently throughout the rest of the dialogue in order to avoid confusing the user. The TaskProgress (P) variable tracks how much progress the user has made completing the experimental task. The CurrentUserGoal (G) variable corresponds to the system's belief about what the user's current goal is. The WhichSelection (W) variable tracks whether the system knows what type of selection criteria the user would like to use to read her messages. The KnowSelectionCriteria (SC) variable tracks whether the system believes it understood either a sender name or a subject name to use to select messages. The NumMatches (M) variable keeps track of how many messages match the user's selection criteria. The Confidence (C) variable is a threshholded variable indicating whether the speech recognizer's confidence that it understood what the user said was above a pre-set threshhold. The Timeout (T) variable represents the system's belief that the user didn't say anything in the allotted time. The Help (H) variable represents the system's belief that the user said *Help*, and leads to the system providing context-specific help messages. The Cancel (L) variable represents the system's belief that the user said *Cancel*, which leads to the system resetting the state to the state before the last user utterance was processed. Thus there are 110,592 possible states used to control the operation of the system, although not all of the states occur.[3]

---

2. In some respects this is driven by implementation requirements since system development and maintenance is impossible without compact state representations.

| U | I | S | R | P | G | M | W | SC | C | T | H | L | Action Choices |
|---|---|---|---|---|---|---|---|---|---|---|---|---|---|
| | | | Operations Variables | | | | | | | | | | Action Choices |
| 0 | 0 | 0 | 0 | 0 | 0 | 0 | 0 | 0 | 0 | 0 | 0 | 0 | AskUserName |
| 1 | 0 | 0 | 0 | 0 | 0 | 0 | 0 | 0 | 1 | 0 | 0 | 0 | SI-Top, MI-Top |
| 1 | SI | 0 | 0 | 0 | 0 | 0 | 0 | 0 | 1 | 0 | 1 | 0 | SI-Top-Help |
| 1 | SI | 0 | 0 | 0 | 0 | 0 | 0 | 0 | 0 | 0 | 0 | 0 | SI-Top-Reject |
| 1 | SI | 0 | 0 | 0 | S | 0 | 0 | 0 | 1 | 0 | 0 | 0 | SS,SB,SCP |
| 1 | SI | 0 | 0 | 0 | R | 0 | 0 | 0 | 1 | 0 | 0 | 0 | AskWS |
| 1 | SI | 0 | 0 | 0 | R | 0 | 0 | 0 | 0 | 0 | 0 | 0 | AskWS-Reject |
| 1 | SI | 0 | 0 | 0 | R | 0 | Snd | 0 | 1 | 0 | 0 | 0 | AskSC |
| 1 | SI | 0 | 0 | 0 | R | 0 | Snd | 0 | 1 | 1 | 0 | 0 | AskSC-TimeOut |
| 1 | SI | 0 | 0 | 0 | R | N>1 | Snd | 1 | 1 | 0 | 0 | 0 | RF,RSO,RCP |
| 1 | SI | 0 | RCP | 0 | R | 1 | Snd | 1 | 1 | 0 | 0 | 0 | ReadMessage |
| 1 | SI | 0 | RCP | 1 | 0 | 0 | 0 | 1 | 0 | 0 | 0 | 0 | SI-Top |
| 1 | MI | 0 | 0 | 0 | 0 | 0 | 0 | 0 | 1 | 0 | 1 | 0 | MI-Top-Help |
| 1 | MI | 0 | 0 | 0 | 0 | 0 | 0 | 0 | 0 | 0 | 0 | 0 | MI-Top-Reject |
| 1 | MI | 0 | 0 | 0 | S | 0 | 0 | 0 | 1 | 0 | 0 | 0 | SS,SB,SCP |
| 1 | MI | SS | 0 | 0 | R | N>1 | Snd | 1 | 1 | 0 | 0 | 0 | RF,RSO,RCP |
| 1 | MI | SS | RF | 0 | R | 1 | Snd | 1 | 1 | 0 | 0 | 0 | ReadMessage |

Figure 7: A portion of ELVIS's operations state machine using the full operations vector to control ELVIS's behavior

In order for the reader to achieve a better understanding of the range of ELVIS's capabilities and the way the operations vector is used, Figure 7 shows a portion of ELVIS's state machine that can generate the sample system and mixed-initiative dialogue interactions in Figures 8 and 9. Each of these figures provides the state representation and the strategy choices made in each state of the sample dialogues. For example, row two of Figure 7 shows that after the system acquires the user's name (KnowUserName (U) = 1) with high confidence (Confidence (C) = 1), that it can explore the system-initiative (SI-Top) or mixed-initiative (MI-Top) strategies. Figure 8 illustrates a dialogue in which the SI strategy was chosen while Figure 9 illustrates a dialogue in which the MI-Top strategy was chosen. Here we discuss in detail how the dialogue in Figure 8 was generated by the state machine in Figure 7.

---

3. For example until the system knows the user name, none of the other variable values change from their initial value.

In Figure 8, the first row shows that Elvis's strategy AskUserName is executed in the initial state of the dialogue where all the operations variables are set to 0. Elvis's utterance E1 is the surface realization of this strategy's execution. Note that according to the state machine in Figure 7, there are no other strategy choices for the initial state of the dialogue. The user responds with her name and the SLU module returns the user's name to the dialogue manager with high confidence (Confidence (C) = 1). The dialogue manager updates the operations variables with KnowUserName(U) = 1 and Confidence (C) = 1, as shown in row two of Figure 8. Now, according to the state machine in Figure 7, there are two choices of strategy, the system-initiative strategy whose initial action is SI-Top and the mixed-initiative strategy whose initial action is MI-Top. Figure 8 illustrates one potential path when the SI-Top strategy is chosen; Elvis's utterance E2 is the realization of the SI-Top strategy. The user responds with the utterance *Help* which is processed by SLU, and the dialogue manager receives as input the information that SLU believes that the user said *Help* (Help (H) = 1) with high confidence (Confidence (C) = 1). The dialogue manager updates the operations variables to reflect the information from SLU as well as the fact that it executed the system-initiative strategy (InitStrat (I) = SI). This results in the operations vector shown adjacent to Elvis's utterance E3. The third row of the state machine in Figure 7 shows that in this state, Elvis has no choice of strategies, so Elvis simply executes the SI-Top-Help strategy, which is realized as utterance E3. The user responds by saying *Read* (utterance U3) and the dialogue manager updates the operations variables with the results of the SLU module saying that it believes that the user said *Read* (Goal (G) = R) with high confidence (Confidence (C) = 1). The state machine in Figure 7 specifies that in this state Elvis should execute the AskWhichSelection (AskWS) strategy, which corresponds to Elvis's utterance E4 in Figure 8. This time, however, when the user responds to the system's query with the word *Sender* (utterance U4), the SLU module is not confident of its understanding (Confidence (C) = 0) and the operations variable is updated with this confidence value. According to the state machine in Figure 7, the strategy that Elvis executes in this state is the AskWS-Reject strategy, a specific rejection strategy designed to be executed in the context of a goal to acquire the value of the WhichSelection operations variable (see the list of operations variables in Figure 6). The realization of the AskWS-Reject strategy is utterance E5. The user responds by saying *Sender* again (utterance U5) and this time the SLU module understands the user's utterance with high confidence. The dialogue manager updates the WhichSelection (W) variable with Snd and the confidence variable, resulting in the operations vector shown adjacent to utterance E6. The state machine in Figure 7 specifies that in this state Elvis should execute the AskSelectionCriterion (AskSC) strategy which is realized as utterance E6. The user says nothing (utterance U6) and the SLU module detects a silence above a pre-set timeout threshhold with high confidence. The dialogue manager updates the operations variables to reflect this observation (Timeout (T) = 1, Confidence (C) = 1). The state machine in Figure 7 specifies that the AskSC-Timeout strategy should be executed in this state; this is a context specific strategy that re-attempts the system's initial goal by providing information

| Operations Variables | | | | | | | | | | | | | Utterance | Strategy Name |
|---|---|---|---|---|---|---|---|---|---|---|---|---|---|---|
| U | I | S | R | P | G | M | W | SC | C | T | H | L | | |
| 0 | 0 | 0 | 0 | 0 | 0 | 0 | 0 | 0 | 0 | 0 | 0 | 0 | E1: Hi, ELVIS here, Who is that? <br> U1: Susan. | AskUserName |
| 1 | 0 | 0 | 0 | 0 | 0 | 0 | 0 | 0 | 1 | 0 | 0 | 0 | E2: Hi Susan, Please wait. I'll get your mail (10 seconds). Okay Susan, You have 5 new or unread messages in your inbox. Say Read, Summarize or say Help for more options. <br> U2: Help. | SI-Top |
| 1 | SI | 0 | 0 | 0 | 0 | 0 | 0 | 0 | 1 | 0 | 1 | 0 | E3: To access your messages, you can say read or summarize. To switch the active messages, say new messages, or all messages. If you want me to repeat a message or a summary, say repeat. You can always say cancel or help. Say Goodbye Elvis, to hang up. <br> U3: Read. | SI-Top-Help |
| 1 | SI | 0 | 0 | 0 | R | 0 | 0 | 0 | 1 | 0 | 0 | 0 | E4: Select by Sender, by Subject or In order of Arrival? <br> U4: Sender | AskWS |
| 1 | SI | 0 | 0 | 0 | R | 0 | 0 | 0 | 0 | 0 | 0 | 0 | E5: I didn't understand you. Do you want to select messages by sender, subject or in order of arrival? <br> U5: Sender | AskWS-Reject |
| 1 | SI | 0 | 0 | 0 | R | 0 | Snd | 0 | 1 | 0 | 0 | 0 | E6: Which Sender? <br> U6: (user says nothing) | AskSC |
| 1 | SI | 0 | 0 | 0 | R | 0 | Snd | 0 | 1 | 1 | 0 | 0 | E7: The senders are, Kim, Noah, Michael, and Owen. Which sender do you want? <br> U7: Kim. | AskSC-TimeOut |
| 1 | SI | 0 | 0 | 0 | R | N | Snd | 1 | 1 | 0 | 0 | 0 | E8: In the messages from Kim, there's 1 message about "Interviewing Antonio" and 1 message about "Meeting Today." To hear the messages, say, "Interviewing Antonio" or "Meeting." <br> U8: Meeting. | RCP |
| 1 | SI | 0 | RCP | 0 | R | 1 | Snd | 1 | 1 | 0 | 0 | 0 | E9: This is a message about meeting today. It says the meeting this morning is at 10:00 in 2d516. See you there, Kim. <br> U9: I'm done here. | ReadMessage |
| 1 | SI | 0 | RCP | 1 | 0 | 0 | 0 | 0 | 1 | 0 | 0 | 0 | E10: You are back to your toplevel inbox. Say Read, Summarize or say Help for more options. | SI-Top |

Figure 8: A System-Initiative Dialogue, completing Task 1.1 in Figure 11, illustrating ELVIS's ability to provide help, and use timeout and confidence information

that is intended to help the user and then re-asking the original query, as realized by utterance E7. The user responds with the name of the sender (utterance U7) which is understood by SLU with high confidence (KnowSelectionCriteria (SC) = 1, Confidence =

1). When Elvis retrieves messages from the mail server matching this selection criteria, multiple matches are found (NumMatches = N, as per the list of operations variables in Figure 6). This time row ten of the state machine in Figure 7 specifies that this state has a choice of dialogue strategies, namely a choice between the Read-First (RF), Read-Summary-Only (RSO) and Read-Choice-Prompt (RCP) strategies illustrated in Figure 5. Elvis randomly chooses to explore the RCP strategy, which is realized as utterance E8. The information the user needs to complete Task 1.1 is then provided by utterance E9 after the user responds in utterance U8 by saying *Meeting* (and SLU understands this with high confidence). The row with utterance E9 in Figure 8 shows the updated operations vector reflecting the fact that the system executed the RCP strategy; the ReadStrat (R) variable is used to enforce the fact that in this implementation of Elvis, once a particular reading, strategy is selected, it is then used consistently throughout the dialogue to avoid confusing the user. In the last exchange of Figure 8, the SLU module's confident understanding of the user's utterance in U9, *I'm done here*, results in resetting the G,M,W, and SC variables and the dialogue manager also updates the variable TaskProg (P) to 1 to reflect progress on the experimental task. Figure 7 shows that, in this state, the system has only one strategy; since the InitStrat variable has been set to SI, the system executes the SI-Top strategy, as realized in this context by utterance E10.

The dialogue in Figure 9 illustrates a potential dialogue with Elvis when the MI-Top strategy is selected rather than the SI-Top strategy after the user name is acquired. The reader may also track the path of this dialogue by utilizing the state machine in Figure 7.

Note that the operations vector that Elvis utilizes is needed to make Elvis a fully operational system that provides all the functionality equired to support users. The dialogues in Figures 8 and 9 also show how Elvis provides:

- Context-Specific Help strategies: illustrated by the strategies SI-Top-Help and MI-Top-Help, and supported by the Help variable.

- Timeout strategies: that the system uses for taking the initiative in each context, triggered by the delay for the user's response going above a time threshhold, as illustrated by the strategy AskSC-TimeOut, and supported by the Timeout variable.

- Rejection strategies: backoff dialogue actions that become more directive to the user when the ASR returns a confidence value lower than its confidence threshhold, as illustrated by strategies AskWS-Reject and MI-Top-Reject, and supported by the Confidence variable.

However, the operations vector state representation needed to control the operation of the system is not necessarily required or even optimal for applying reinforcement learning (Barto et al., 1995; Sutton & Barto, 1998). Sometimes it may be advantageous to aggregrate states for the purpose of applying reinforcement learning, even if it is not possible to guarantee that the state representation obeys the Markov property (Sutton & Barto, 1998). Note that in many of the states defined in Figure 7 alternate strategies are not explored. For example, the Confidence variable is used by the system to determine whether it should apply a backoff rejection strategy, but we do not vary the rejection strategies. The incentive for aggregrating states by ignoring state variables that are irrelevant to learning

| Operations Variables | | | | | | | | | | | | Utterance | Strategy Name |
|---|---|---|---|---|---|---|---|---|---|---|---|---|---|
| U | I | S | R | P | G | M | W | SC | C | T | H | L | | |
| 0 | 0 | 0 | 0 | 0 | 0 | 0 | 0 | 0 | 0 | 0 | 0 | 0 | E1: Hi, ELVIS here, Who is that?<br>U1: Susan. | AskUserName |
| 1 | 0 | 0 | 0 | 0 | 0 | 0 | 0 | 0 | 1 | 0 | 0 | 0 | E2: Hi Susan, Please wait. I'll get your mail (*delay of about 10 seconds*).<br>Okay Susan, I've got your mail.<br>U2: Help. | MI-Top |
| 1 | MI | 0 | 0 | 0 | 0 | 0 | 0 | 0 | 1 | 0 | 1 | 0 | E3: You can read or summarize messages using values from the sender or the subject field. For example, you can say, read the message from Lee about the meeting, or, read my messages from Kim, or, summarize my messages. To access messages in the current folder, you can also say next, previous, first, or last. If you want me to repeat a message or a summary, say repeat. You can always say cancel or help. To exit the current folder say, I'm done here. To hangup, say, Goodbye Elvis .<br>U3: Summarize my messages. | MI-Top-Help |
| 1 | MI | 0 | 0 | 0 | 0 | 0 | 0 | 0 | 0 | 0 | 0 | 0 | E4: I didn't understand you. What do you want to do with your messages?<br>U4: Summarize them. | MI-Top-Reject |
| 1 | MI | 0 | 0 | S | 0 | 0 | 0 | 0 | 1 | 0 | 0 | 0 | E5: In your top level inbox, there's 2 messages from Kim, 1 message from Noah, 1 message from Michael, and 1 message from Owen.<br>U5: Read my messages from Kim | SS |
| 1 | MI | SS | 0 | 0 | R | N | Snd | 1 | 1 | 0 | 0 | 0 | E6: In the messages from Kim, there's one message about Interviewing Antonio and one message about Meeting Today. The first message says (*user barges in*)<br>U6: Next message | RF |
| 1 | MI | SS | RF | 1 | R | 1 | Snd | 1 | 1 | 0 | 0 | 0 | E7: Next message. This is a message about meeting today. It says the meeting this morning is at 10:00 in 2d516. See you there, Kim. | ReadMessage |

Figure 9: A Mixed Initiative Dialogue completing Task 1.1 in Figure 11, illustrating ELVIS's ability to provide help, and use timeout and confidence information

is a reduction in the state space size; this means that fewer dialogue samples are needed to collect a large enough sample of state/action pairs for the purpose of applying reinforcement learning. From this perspective, our goal is to aggregate the state space in such a way as to only distinguish states where different dialogue strategies are explored.

However, there is an additional constraint on state aggregation. Reinforcement learning[4] backs up rewards received in the final states of the dialogue $s_f$ to earlier states $s_i$ where strategy choices were explored. However the algorithm can only distinguish strategy choices when the trajectory between $s_i$ and $s_f$ are distinct for each strategy choice. In other words, if two actions at some point lead to the same state, then without local reward, the Q-values of these two actions will be equal.

| UserName (U) | Init (I) | TaskProg (P) | UserGoal (G) |
|---|---|---|---|
| 0,1 | 0,SI,MI | 0,1,2, | 0,R,S |

Figure 10: Reinforcement Learning State Variables and Values

Figure 10 specifies the subset of the state variables given in Figure 6 that we developed to represent the state space for the purpose of applying reinforcement learning. The combination of these state variables is very compact, but provides distinct trajectories for different strategy choices. The reduced state space has only 18 states, but supports dialogue optimization over a policy space of $2 * 3^{12} = 1062882$ different policies. All of the policies are prima facie candidates for optimal policies in that they can all support human users in completing a set of experimental email tasks.

## 3. Experimental Design

Experimental dialogues for both the training and testing phase were collected via experiments in which human users interacted with Elvis to complete three representative application tasks that required them to access email messages in three different email inboxes. We collected data from 73 users performing three tasks (219 dialogues) for training Elvis, and then tested the learned policy against a corpus from six users performing the same three tasks (18 dialogues).

Instructions to the users were given on a set of web pages, with one page for each experimental dialogue. The web page for each dialogue also contained a brief general description of the functionality of the system, a list of hints for talking to the system, a description of the tasks that the user was supposed to complete, and information on how to call Elvis. Each page also contained a form for specifying information acquired from Elvis during the dialogue, and a survey, to be filled out after task completion, designed to probe the user's satisfaction with Elvis. Users read the instructions in their offices before calling Elvis from their office phone.

Each of the three calls to Elvis was made in sequence, and each conversation consisted of two task scenarios where the system and the user exchanged information about criteria for selecting messages and information within the message. The tasks are given in Figure 11, where, e.g., Task 1.1 and Task 1.2 were done in the same conversation with Elvis. The motivation for asking the caller to complete multiple tasks in a call was to create subdialogue structure in the experimental dialogues (Litman, 1985; Grosz & Sidner, 1986).

---

4. When applied without local rewards.

- Task 1.1: You are working at home in the morning and plan to go directly to a meeting when you go into work. Kim said she would send you a message telling you where and when the meeting is. Find out **the Meeting Time** and **the Meeting Place**.

- Task 1.2: The second task involves finding information in a different message. Yesterday evening, you had told Lee you might want to call him this morning. Lee said he would send you a message telling you where to reach him. Find out **Lee's Phone Number**.

- Task 2.1: When you got into work, you went directly to a meeting. Since some people were late, you've decided to call ELVIS to check your mail to see what other meetings may have been scheduled. Find out the **day, place** and **time** of any scheduled meetings.

- Task 2.2: The second task involves finding information in a different message. Find out if you need to call anyone. If so, find out **the number** to call.

- Task 3.1: You are expecting a message telling you when the Discourse Discussion Group can meet. Find out the **place** and **time** of the meeting.

- Task 3.2: The second task involves finding information in a different message. Your secretary has taken a phone call for you and left you a message. Find out **who called** and **where** you can reach them.

Figure 11: Sample Task Scenarios

- **Dialogue Efficiency Metrics**: elapsed time, system turns, user turns
- **Dialogue Quality Metrics** mean recognition score, timeouts, rejections, helps, cancels, bargeins, timeout%, rejection%, help%, cancel%, bargein%
- **Task Success Metrics**: task completion as per survey
- **User Satisfaction**: the sum of TTS Performance, ASR Performance, Task Ease, Interaction Pace, User Expertise, System Response, Expected Behavior, Comparable Interface, Future Use.

Figure 12: Metrics collected for spoken dialogues.

We collect a number of of different measures for each dialogue via four different methods: (1) All of the dialogues are recorded; (2) The dialogue manager logs each state that the system enters and the dialogue strategy that ELVIS selects in that state; (3) The dialogue manager logs information for calculating a number of dialogue quality and dialogue efficiency metrics summarized in Figure 12 and described in more detail below; and (4) At the end of the dialogue, users fill out web page forms to support the calculation of task success and user satisfaction measures. We explain below how we use these measures in the PARADISE framework and in reinforcement learning.

The **dialogue efficiency** metrics were calculated from the dialogue recordings and the system logs. The length of the recording was used to calculate the elapsed time in seconds (**ET**) from the beginning to the end of the interaction. Measures for the number of **System Turns**, and the number of **User Turns**, were calculated on the basis of the system logging everything it said and everything it heard the user say.

The **dialogue quality** measures were derived from the recordings, the system logs and hand-labeling. A number of system behaviors that affect the quality of the resulting dialogue were automatically logged. These included the number of timeout prompts (**timeouts**) played when the user didn't respond as quickly as expected, the number of recognizer rejections (**rejects**) where the system's confidence in its understanding was low and it said something like *I'm sorry I didn't understand you*. User behaviors that the system perceived that might affect the dialogue quality were also logged: these included the number of times the system played one of its context specific help messages because it believed that the user had said *Help* (**helps**), and the number of times the system reset the context and returned to an earlier state because it believed that the user had said *Cancel* (**cancels**). The recordings were used to check whether users barged in on system utterances, and these were labeled on a per-state basis (**bargeins**).

Another measure of dialogue quality was recognizer performance over the whole dialogue, calculated in terms of concept accuracy. The recording of the user's utterance was compared with the logged recognition result to calculate a concept accuracy measure for each utterance by hand. Concept accuracy is a measure of semantic understanding by the system, rather than word for word understanding. For example, the utterance *Read my messages from Kim* contains two concepts, the *read* function, and the *sender:kim* selection criterion. If the system understood only that the user said *Read*, then concept accuracy would be 0.5. Mean concept accuracy was then calculated over the whole dialogue and used, in conjunction with ASR rejections, to compute a Mean Recognition Score (**MRS**) for the dialogue.

Because our goal is to generate models of performance that will generalize across systems and tasks, we also thought it important to introduce metrics that are likely to generalize. All of the efficiency metrics seemed unlikely to generalize since, e.g., the **elapsed time** to complete a task depends on how difficult the task is. Other research suggested that the dialogue quality metrics were more likely to generalize (Litman, Walker, & Kearns, 1999), but we thought that the raw counts were likely to be task specific. Thus we normalized the dialogue quality metrics by dividing the raw counts by the total number of utterances in the dialogue. This resulted in the **timeout%, rejection%, help%, cancel%**, and **bargein%** metrics.

The web page forms are the basis for calculating Task Success and User Satisfaction measures. Users reported their perceptions as to whether they had completed the task (**Comp**).[5] They also had to provide objective evidence that they had in fact completed the task by filling in a form with the information that they had acquired from ELVIS.[6]

---

5. *Yes,No* responses are converted to *1,0*.
6. This supports an alternative way of calculating **Task Success** objectively by using the Kappa statistic to compare the information that the users filled in with a key for the task (Walker et al., 1997a). However some of our earlier results indicated that the user's *perception* of task success was a better predictor of overall satisfaction, so here we simply use perceived task success as measured by **Comp**.

- Was ELVIS easy to understand in this conversation? (**TTS Performance**)
- In this conversation, did ELVIS understand what you said? (**ASR Performance**)
- In this conversation, was it easy to find the message you wanted? (**Task Ease**)
- Was the pace of interaction with ELVIS appropriate in this conversation? (**Interaction Pace**)
- In this conversation, did you know what you could say at each point of the dialogue? (**User Expertise**)
- How often was ELVIS sluggish and slow to reply to you in this conversation? (**System Response**)
- Did ELVIS work the way you expected him to in this conversation? (**Expected Behavior**)
- In this conversation, how did ELVIS's voice interface compare to the touch-tone interface to voice mail? (**Comparable Interface**)
- From your current experience with using ELVIS to get your email, do you think you'd use ELVIS regularly to access your mail when you are away from your desk? (**Future Use**)

Figure 13: User Satisfaction Survey

In order to calculate User Satisfaction, users were asked to evaluate the system's performance with the user satisfaction survey in Figure 13. Some of the question responses were on a five point Likert scale and some simply required *yes, no* or *yes, no, maybe* responses. The survey questions probed a number of different aspects of the users' perceptions of their interaction with ELVIS in order to focus the user on the task of rating the system, as in (Shriberg et al., 1992; Jack, Foster, & Stentiford, 1992; Love, Dutton, Foster, Jack, & Stentiford, 1994). Each multiple choice survey response was mapped into the range of 1 to 5. Then the values for all the responses were summed, resulting in a **User Satisfaction** measure for each dialogue with a possible range of 8 to 40.

## 4. Training and Testing an Optimized Dialogue Strategy

Given the experimental training data, we first apply PARADISE to estimate a performance function for ELVIS as a linear combination of the metrics described above. We apply the performance function to each dialogue in the training corpus to estimate a utility for the final state of the dialogue and then apply Q-learning using this utility. Finally we test the learned policy against a new population of users.

### 4.1 PARADISE Performance Modeling

The first step in developing a performance model for spoken dialogue systems was the specification of the causal model of performance illustrated in Figure 14 (Walker et al., 1997a). According to this model, the system's primary objective is to maximize user satisfaction.

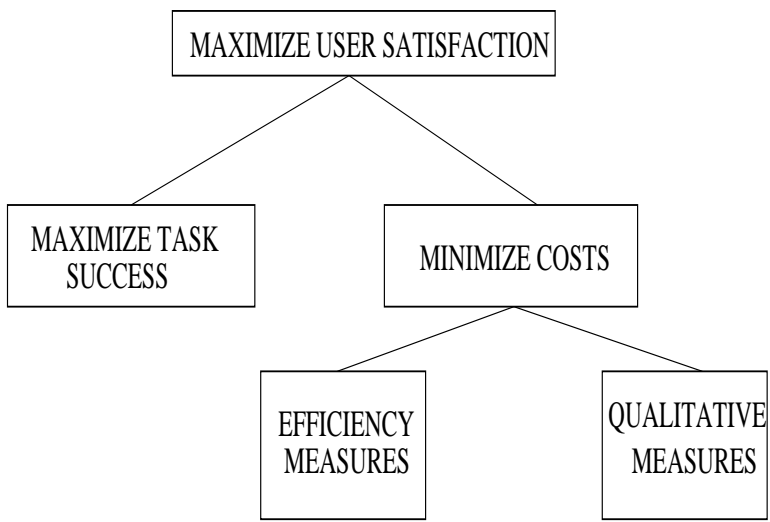

Figure 14: PARADISE's structure of objectives for spoken dialogue performance.

Task success and various costs that can be associated with the interaction are both contributors to user satisfaction. Task success can be measured quantitatively in a number of ways: it could be represented by a continuous variable representing quality of solution or by a boolean variable representing binary task completion. Dialogue costs are of two types: dialogue efficiency and quality. Efficiency costs are measures of the system's efficiency in helping the user complete the task, such as the number of utterances to completion of the dialogue. Dialogue quality costs are intended to capture other aspects of the system that may have strong effects on user's perception of the system, such as the number of times the user had to repeat an utterance in order to make the system understand the utterance.

Given this model, a performance metric for a dialogue system can be estimated from experimental data by applying multivariate linear regression with user satisfaction as the dependent variable and task success, dialogue quality, and dialogue efficiency measures as independent variables.[7] A stepwise linear regression on the training data over the measures discussed above, showed that **Comp**, **MRS**, **BargeIn%** and **Rejection%** were significant contributors to **User Satisfaction**, accounting for 39% of the variance in R-Squared (F $(4,192)=30.7$, $p <.0001$).[8]

$$\text{Performance} = .27 * Comp + .54 * MRS - .09 * BargeIn\% + .15 * Rejection\%$$

We tested how well this performance function should generalize to unseen test dialogues with tenfold cross-validation, by randomly sampling 90% of the training dialogues and testing the goodness of fit of the performance model on the remaining 10% of the dialogues in the training set. The average $R^2$ for the training set was 37% with a standard error of .005, while the average $R^2$ for the held-out 10% of the dialogues was 38% with a standard error of .06. Since the average $R^2$ for the test set is statistically indistinguishable from the training set, we assume that the performance model will generalize to new ELVIS dialogues.

---

7. One advantage of this approach is that once the performance function is derived, it is no longer necessary to collect user satisfaction reports from users, which opens up the possibility of estimating the reward function from fully automatic measures. This latter possibility might also be useful for online calculation of the reward function or for calculating a local reward.
8. We normalize the metrics before doing the regression so that the magnitude of the coefficients directly indicates the contribution of that factor to User Satisfaction (Cohen, 1995; Walker et al., 1997a).

## 4.2 Training an Optimized Policy

Given the learned performance function described above, we apply this function to the measures logged for each dialogue $D_i$, thereby replacing a range of measures with a single performance value $P_i$, which is used as the utility (reward) for the final state of each dialogue.[9] We then apply reinforcement learning with $P_i$ as the utility of the final state of the dialogue $D_i$ (Bellman, 1957; Sutton, 1991; Tesauro, 1992; Russell & Norvig, 1995; Watkins, 1989). The utility of doing action $a$ in state $S_i$, $U(a, S_i)$ (its Q-value), can be calculated in terms of the utility of a successor state $S_j$, by obeying the recursive equation:

$$U(a, S_i) = R(a, S_i) + \sum_j M_{ij}^a \max_{a'} U(a', S_j)$$

where $R(a, S_i)$ is the immediate reward received for doing action $a$ in $S_i$, $a$ is a strategy from a finite set of strategies $A$ that are admissable in state $S_i$, and $M_{ij}^a$ is the probability of reaching state $S_j$ if strategy $a$ is selected in state $S_i$. In the experiments reported here, the reward associated with each state, $R(S_i)$, is zero. In addition, since reliable a priori prediction of a user action in a particular state is not possible (for example the user may say *Help* or the speech recognizer may fail to understand the user), the state transition model $M_{ij}^a$ is estimated from the logged state-strategy history for the dialogue.

The utility values can be estimated to within a desired threshold using value iteration, which updates the estimate of $U(a, S_i)$, based on updated utility estimates for neighboring states, so that the equation above becomes:

$$U_{n+1}(a, S_i) = R(S_i) + \sum_j M_{ij}^a \max_{a'} U_n(a', S_j)$$

where $U_n(a, S_i)$ is the utility estimate for doing $a$ in state $S_i$ after $n$ iterations (Sutton & Barto, 1998) pp. 101. Value iteration stops when the difference between $U_n(a, S_i)$ and $U_{n+1}(a, S_i)$ is below a threshold, and utility values have been associated with states where strategy selections were made.[10] Once value iteration is completed the optimal policy is obtained by selecting the action with the maximal Q-value in each dialogue state.

Figure 15 enumerates the subset of the states in the aggregrated state space used for reinforcement learning and potential actions defining the policy space. The strategy with the greatest Q-value in each state after training is indicated by **boldface** in Figure 15. This optimized policy will then be tested as a fixed policy in the operation of ELVIS. In all the states of the task, the System-Initiative strategy in Figure 2 is predicted to be the optimal initiative strategy, and the Read-First strategy in Figure 5 is predicted to have the best performance of the Read strategies. As Figure 15 shows, the learned strategy for summarization varies according to the state of the task. The different summarization strategies were illustrated in Figure 4. The policy that is learned is to use the Summarize-Both strategy at the beginning of the dialogue (when TaskProg = 0), and then to switch to using the Summarize-System strategy at later phases of the dialogue. This strategy makes sense in terms of giving the user complete information about all the messages in her inbox at the beginning of the dialogue.

---

9. Each dialogue is treated as having a unique final state.
10. After experimenting with various threshholds, we used a threshold of 5% of the performance range of the dialogues.

| State Variables | | | | Strategy Choices |
|---|---|---|---|---|
| U | I | P | G | |
| 0 | 0 | 0 | 0 | AskUserName |
| 1 | 0 | 0 | 0 | **SI-Top**, MI-Top |
| 1 | SI | 0 | S | SS,**SB**,SCP |
| 1 | SI | 0 | R | **RF**,RSO,RCP |
| 1 | SI | 1 | 0 | SI-Top |
| 1 | SI | 1 | S | **SS**,SB,SCP |
| 1 | SI | 1 | R | **RF**,RSO,RCP |
| 1 | SI | 2 | 0 | SI-Top |
| 1 | SI | 2 | S | **SS**,SB,SCP |
| 1 | SI | 2 | R | **RF**,RSO,RCP |
| 1 | MI | 0 | S | SS,**SB**,SCP |
| 1 | MI | 0 | R | **RF**,RSO,RCP |
| 1 | MI | 1 | 0 | MI-Top |
| 1 | MI | 1 | S | **SS**,SB,SCP |
| 1 | MI | 1 | R | **RF**,RSO,RCP |
| 1 | MI | 2 | 0 | MI-Top |
| 1 | MI | 2 | S | **SS**,SB,SCP |
| 1 | MI | 2 | R | **RF**,RSO,RCP |

Figure 15: The subset of the state space that defines the policy class explored in our experiments. The learned policy is indicated by boldface.

### 4.3 Testing an Optimized Policy

We first constructed a deterministic version of ELVIS that implemented the learned policy as discussed above, with one variation. The variation was based on the fact that the decision on whether to use the Summarize-Both or Summarize-System summarization strategy was conditioned on the value of the TaskProg variable. However, we intended to utilize the optimized version of the system in situations where we would not have access to the TaskProg variable, namely situations where the task that the user was attempting to perform were not under the control of the experimenter. When we examined the Q-values for the summarization strategies over the course of the dialogue, we found that the Summarize-System strategy had the greatest average Q-value, being strongly preferred to the Summarize-Both strategy except in the initial phase of the dialogue, where the Q-value for the Summarize-Both was only slightly greater. Thus we implemented the learned policy (see Figure 15), with the exception that the Summarize-System strategy was used throughout the dialogue.[11]

In terms of the operations state machine in Figure 7, implementation of the learned policy means that the choices between the SI-Top and MI-Top strategies are replaced by the SI-Top strategy, choices between the different read strategies in different states are replaced by the Read-First (RF) strategy and choices between the different summarization strategies in different states are replaced by the Summarize-System (SS) strategy.

We then tested this policy on six new users who had never used ELVIS before. These users conversed with ELVIS to perform the same set of six email tasks that were used in the training phase, as described in Figure 10 above. In addition, identical performance measures were collected for each testing dialogue and training dialogue. Overall performance measures for the training and test dialogues are given in Table 1, with the training data split in terms of System-Initiative, Mixed-Initiative and overall means. The table shows that all versions of ELVIS have high levels of task completion, which is important for testing the utility of reinforcement learning. Statistical analysis of these results indicated a statistically significant increase in User Satisfaction from training to test (F= = 4.07 p = .047).

## 5. Discussion and Future Work

This paper proposes a novel method by which a dialogue system can learn to choose an optimal dialogue strategy and tests it in experiments with ELVIS, a dialogue system that supports access to email by phone, with strategies for initiative, and for reading and summarizing messages. We reported experiments in which ELVIS learned that the System-Initiative strategy has higher utility than the Mixed-Initiative strategy, that Read-First is the best read strategy, and that Summarize-System is generally the best summary strategy. We then tested the policy that ELVIS learned on a new set of users performing the same set of tasks and showed that the learned policy resulted in a statistically significant increase in user satisfaction in the test set of dialogues.

Previous work has also treated a system's choice of dialogue strategy as a stochastic optimization problem (Walker, 1993; Biermann & Long, 1996; Levin & Pieraccini, 1997; Levin et al., 1997). To our knowledge, Walker (1993) first proposed that reinforcement learning algorithms could be applied to dialogue strategy selection. In simulation experiments reported by Walker (1993, 1996), dialogues between two agents in an artificial world were used to test which dialogue strategies were optimal under various conditions. These experiments varied: (1) the dialogue agent's resource bounds; and (2) the performance function used to assess the agent's performance. The experiments showed that strategies that were not optimal under one set of assumptions about the performance function could be highly efficacious when the performance function reflected the fact that the dialogue agent was resource bounded. Walker (1993) suggested that the optimal dialogue strategy could be learned via reinforcement learning, if an appropriate performance function could be determined, and described an experiment using genetic algorithms to learn an optimal dialogue strategy. In subsequent work, utilized here, the PARADISE model was proposed as a way to learn an appropriate performance function (Walker et al., 1997a). In addition, related work utilizing ELVIS, that varied the reward function, and applied other reinforcement learning algorithms, was carried out by Fromer (Fromer, 1998).

---

11. Obviously this choice of the strategy to test risked testing a non-optimal policy. An alternative that we would like to try in future work is to utilize only the SummStrat state variable from the operations vector in the state representation for reinforcement learning and simply distinguish states where no summarize strategy has been selected (no summary has been produced) and states where at least one summary has been produced. If the analysis about dialogue phase carries through, then the policy that should be learned is to use the Summarize-Both strategy for the first summary in a dialogue and then afterwards use the Summarize-System strategy.

| Measure | Train SI | Train MI | Overall Train | Test |
|---|---|---|---|---|
| Comp | .87 | .80 | .85 | .94 |
| User Turns | 21.5 | 17.0 | 20.0 | 25.8 |
| System Turns | 24.2 | 21.2 | 23.1 | 29.2 |
| Elapsed time (sec) | 339.14 | 296.18 | 311.56 | 368.5 |
| Mean recognition score | .88 | .72 | .82 | .81 |
| TimeOuts | 2.7 | 4.2 | 3.0 | 3.3 |
| TimeOut% | .11 | .19 | .13 | .11 |
| Cancs | .34 | .02 | .26 | .00 |
| Canc% | .02 | .00 | .01 | .00 |
| Help Requests | .67 | .92 | 0.66 | 1.11 |
| Help% | .03 | .05 | .03 | .04 |
| BargeIns | 3.6 | 3.6 | 3.7 | 7.8 |
| BargeIn% | .08 | .09 | .18 | .30 |
| Rejects | .9 | 1.6 | 1.1 | 1.4 |
| Reject% | .04 | .08 | .05 | .05 |
| User satisfaction | 28.9 | 25.0 | 27.5 | 31.7 |

Table 1: Performance measure means per dialogue for Training and Testing Dialogues. SI = System-Initiative, MI = Mixed-Initiative

Biermann and Long (1996), proposed the use of similar techniques in the context of learning optimal dialogue strategies for a multi-modal dialogue tutor. The goal of the tutor was to instruct students taking their first programming class and the tutor interacted with the students by highlighting parts of their code and printing text on the screen telling them what was wrong with their program. Biermann and Long describe a planned experiment in which the system would vary its instructional style, and the system's reward would be the amount of time between the system's instructions and the student's response. This reward function was based on the assumption that a delayed response suggested a greater cognitive load for the student, and that cognitive load should be minimized in an instructional setting.

Levin and colleagues also proposed treating dialogue systems as Markov Decision Processes and suggested that system designers could determine what an appropriate objective function might be (Levin et al., 1997; Levin & Pieraccini, 1997). They carried out a series of experiments in which a simulated user interacted with an implemented spoken dialogue system for travel planning by exchanging messages at the semantic meaning level. They showed that the system could learn strategy choices at the level of database interaction, e.g., that the system should not query the database until it had determined many of the constraints necessary in order to find flights that matched the user's goals.

Stochastic optimization techniques have also been applied to similar problems in text-based dialogue interaction and graphical user interfaces. Mellish and colleagues applied stochastic optimization to the problem of determining the content and structure of the system's utterances in the ILEX system, an interactive museum tour guide (Mellish et al., 1998). This work was not tested against a user population and the performance (reward) measure was based on heuristics about good text plans formulated by experts. Christensen and colleagues applied genetic algorithms to the design of a graphical user interface for an automated teller machine. The goal was to automatically learn the best layout of a sequence of interaction screens for intracting with a user (Christensen, Marks, & Shieber, 1994). In this work, as in that of Levin and colleagues, the user population was simulated.

Here, the method for optimizing dialogue strategy selection was illustrated by evaluating strategies for managing initiative and for information presentation by interaction with human callers. We applied the PARADISE performance model to derive an empirically motivated performance function, that combines both subjective user preferences and objective system performance measures into a single function. It would have been impossible to predict a priori which dialogue factors influence the usability of a dialogue system, and to what degree. Our performance equation shows that task success and dialogue quality measures are the primary contributors to system performance. Furthermore, in contrast to assuming an a priori model, we use the dialogues from real user-system interactions to provide realistic estimates of $M_{ij}^a$, the state transition model used by the learning algorithm. It is impossible to predict a priori the transition frequencies, given the imperfect nature of spoken language understanding, and the unpredictability of user behavior.

The use of this method introduces several open issues and possible areas for future work. First, the results of the learning algorithm are dependent on the representation of the state space. In spoken dialogue systems, the system designers construct the state space and decide what state variables the system needs to monitor, whereas in other applications of reinforcement learning (e.g. backgammon), the state space is pre-defined. In the experiments reported here, we fixed the state representation and carried out experiments on a particular state representation. However in future work we hope to be able to learn which aspects of the state history should be represented using similar techniques to those described in (Langkilde, Walker, Wright, Gorin, & Litman, 1999). For example, it may be beneficial for the system to represent additional state variables representing more of the dialogue history, in order for ELVIS to be able to learn dialogue strategies that reflect those aspects of the dialogue history.

Second, in advance of actually running experiments, it is not clear how much experience a system will need to determine which strategy is better. In the experiments reported here, we were able to show an improvement for a policy that had converged on the initiative and read strategies but had not yet converged on the appropriate summarization strategy. It is possible that if our local rewards had been nonzero that the optimal policy could have been learned from less training data. In future work, we hope to explore the interaction of training set size and the use of a local reward.

Third, our experimental data is based on fixing particular experimental parameters. All of the experiments are based on short-term interactions with novice users, but we might expect that users of an email system would engage in many interactions with the same system, and that preferences for system interaction strategies could change over time with user expertise. This means that the performance function might change over time. We also used a fixed set of tasks that were representative of the domain, but it is possible that some aspects of the policies we learned might be sensitive to our experimental tasks. Another limitation is that the experiments were carried out in a scenario where each email folder only had a small number of messages: the strategies tested here might not be optimal when an email folder contains hundreds of messages.

Fourth, the optimal strategy is potentially dependent on various system parameters. For example, the ReadFirst strategy takes the initiative to read a message, which might result in messages being read that the user wasn't interested in, but since the user can barge-in on system utterances, there is little overhead with taking this decision. If the system did not support barge-in, our results might have been different.

Fifth, the learned policy depends on the reward function. For example, since Elvis is a fully functional system, users can complete the experimental task with any version of the system using any of the strategies that we explored. This means that if we had used task completion as the reward function, reinforcement learning would have predicted that there were no differences between the different strategies. On the other hand, by using the paradise performance function, we utilized a reward function that was fit to our data and Elvis's performance, and we have some evidence that this reward function may generalize to other systems (Walker, Kamm, & Litman, 2000).

Sixth, the experiments that we report here are limited in the way that they demonstrate the utility of reinforcement learning for dialogue strategy optimization. A more traditional way of selecting the best dialogue strategies would be with experiments which treated dialogue strategy selection as a factor, and standard statistical hypothesis testing would be used to compare the performance of different strategies. The scale of the experiment here is small enough that it is imaginable that the space of policies could possibly have been tested in the more traditional way. However, the primary goal of the experiments reported here was simply to test the feasibility of these methods, which required working out in detail many of the issues of state and strategy representation discussed above. Now that many of these details have been worked out, the methods presented here can be applied to much more complex dialogue strategy optimization problems, such as varying the initiative depending on the dialogue state (Chu-Carroll & Brown, 1997; Webber & Joshi, 1982), or exploring combinations of strategies for information presentation, summarization (Sparck-Jones, 1999), error recovery (Hirschman & Pao, 1993), database query (Levin et al., 1997), cooperative responses (Joshi, Webber, & Weischedel, 1986; Finin et al., 1986; Chu-Carroll & Carberry, 1994), and content selection for generation (McKeown, 1985; Kittredge, Korelsky, & Rambow, 1991), *inter alia*.

Finally, the learning algorithm that we report here is an *off-line* algorithm, i.e. Elvis collects a set of dialogues and then decides on an optimal strategy as a result. In contrast, it should be possible for Elvis to learn *on-line*, during the course of a dialogue, once methods are developed for the performance function to be automatically calculated or approximated.

Our primary goal with the experiments reported here was to explore the application of reinforcement learning to spoken dialogue systems and to identify open issues such as those discussed above. In current work, we are exploring these issues in several ways. We have codified the notion of a state estimator so that we can systematically vary the state representation in order to explore the effect of the state representation on the value function and the optimal policy (Singh, Kearns, Litman, & Walker, 1999). We are also in the process of using reinforcement learning to conduct a set of experiments on a spoken dialogue system for accessing information about activities in New Jersey. In these experiments we explore a number of different reward functions and also explore a much broader range of strategies for user initiative, for reprompting the user, and for confirming the system's understanding.

## 6. Acknowledgements

I received many useful questions and comments on this research when I presented some initial results at an invited talk given at AAAI 1997 in Providence, R.I. The design and implementation of the basic functionality of Elvis was done in collaboration with J. Fromer, G. DiFabbrizio, D. Hindle and C. Mestel. Initial experiments on reinforcement learning with Elvis were done in collaboration with J. Fromer and S. Narayanan. This work has also benefited from discussions with W. Eckert, C. Kamm, M. Kearns, E. Levin, D. Litman, D. McAllester, R. Pieraccini, R. Sutton, and S. Singh. Special thanks are to S. Whittaker, J. Wiebe and four reviewers for detailed comments on earlier versions of this manuscript.
　v